\pdfoutput=1

\documentclass[11pt]{article}
\usepackage{float}
\usepackage[preprint]{acl}

\usepackage{times}
\usepackage{latexsym}
\usepackage{booktabs}
\usepackage{arydshln}

\usepackage[T1]{fontenc}

\usepackage[utf8]{inputenc}

\usepackage{microtype}

\usepackage{inconsolata}

\usepackage{graphicx}
\usepackage{enumitem}
\setlist[itemize]{noitemsep, topsep=0pt, leftmargin=11pt}
\setlist[enumerate]{noitemsep, topsep=0pt, leftmargin=11pt}

\title{Bridging the Creativity Understanding Gap: Small-Scale Human Alignment Enables Expert-Level Humor Ranking in LLMs}

\author{
\textbf{Kuan Lok Zhou\thanks{\,\,\,Equal contribution, random draw.}$^{,1}$},
\textbf{Jiayi Chen$^{*,1}$},
\textbf{Siddharth Suresh$^{1}$}, \textbf{Reuben Narad$^{2}$}, 
\textbf{Timothy T. Rogers$^{1}$}, \\
\textbf{Lalit K Jain$^{2}$, Robert D Nowak$^{1}$, Bob Mankoff$^{3}$, Jifan Zhang$^{1}$} \\
\\
\textsuperscript{1}University of Wisconsin-Madison,
\textsuperscript{2}University of Washington, Seattle,\\
\textsuperscript{3}Air Mail and Cartoon Collections
}

\begin{document}
\maketitle

\begin{abstract}
Large Language Models (LLMs) have shown significant limitations in understanding creative content, as demonstrated by \citet{hessel-etal-2023-androids}'s influential work on the New Yorker Cartoon Caption Contest (NYCCC). Their study exposed a substantial gap between LLMs and humans in humor comprehension, establishing that understanding and evaluating
creative content is key challenge in AI development. We revisit this challenge by decomposing humor understanding into three components and systematically improve each: enhancing visual understanding through improved annotation, utilizing LLM-generated humor reasoning and explanations, and implementing targeted alignment with human preference data. Our refined approach achieves 82.4\% accuracy in caption ranking, singificantly improving upon the previous 67\% benchmark and matching the performance of world-renowned human experts in this domain. Notably, while attempts to mimic subgroup preferences through various persona prompts showed minimal impact, model finetuning with crowd preferences proved remarkably effective. These findings reveal that LLM limitations in creative judgment can be effectively addressed through focused alignment to specific subgroups and individuals. Lastly, we propose the position that achieving artificial general intelligence necessitates systematic collection of human preference data across creative domains. We advocate that just as human creativity is deeply influenced by individual and cultural preferences, training LLMs with diverse human preference data may be essential for developing true creative understanding.
\end{abstract}

\section{Introduction}
\textcolor{red}{\textbf{Warning: this paper contains potentially offensive content due to the nature of humor.}}

\noindent The emergence of Large Language Models (LLMs) has revolutionized many domains of artificial intelligence, yet their ability to understand and evaluate creative content remains notably limited. This limitation is particularly evident in humor comprehension, as demonstrated by \citet{hessel-etal-2023-androids}'s seminal work on the New Yorker Cartoon Caption Contest (NYCCC). Their study, which earned the best paper award at ACL 2023, exposed a substantial gap between LLMs and human performance in ranking humorous captions, establishing creative understanding as a key challenge in AI systems.

\begin{figure*}[ht]
\centering
\includegraphics[width=\textwidth]{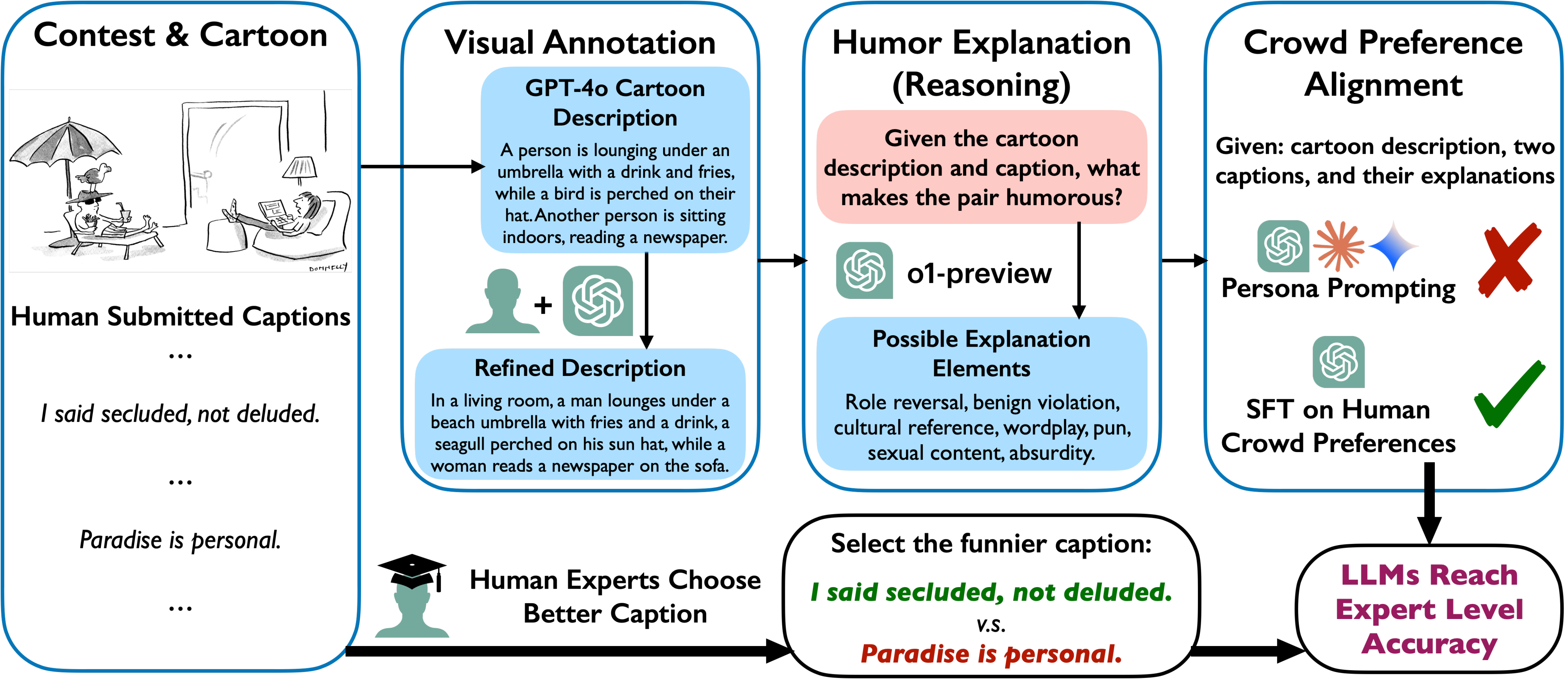}
\caption{Our work improve over state-of-art caption ranking through a three-stage process. With multimodel LLM assistance, we manually fix visual understanding and cartoon description flaws. Our framework also incorporates o1 reasoning capabilities in explaining a joke, before utilizing two different alignment methods to align an LLM preferences with the human preferences from the NYYCC. Our experiments demonstrate that we are achieving human expert level accuracy in this caption ranking task.}
\label{fig:schematic}
\end{figure*}

We revisit this challenge by decomposing humor understanding into three components: visual understanding, cartoon-caption reasoning, and alignment with human preferences as demonstrated in Figure~\ref{fig:schematic}. Through improved visual annotations and LLM-generated explanations, we significantly enhanced both visual understanding and cartoon-caption reasoning. However, the most critical and challenging component proved to be alignment with human preferences.

Our work reveals an intriguing paradoxical finding in this alignment challenge: while LLMs can now generate sophisticated and accurate explanations about why captions are humorous, they still struggle with the seemingly easier task of ranking pairs of captions. Our attempts to bridge this gap through various persona-based prompting techniques showed minimal impact, suggesting a fundamental limitation in how LLMs understand human preferences. The breakthrough came through explicit finetuning on human preference data from the caption contest crowd. Combined with the other improvements mentioned above, we dramatically increased our ranking performance from 67\% to 82.4\% accuracy, matching or exceeding the performance of human experts. This success extends to an even more challenging variant of the task where the crowd-averaged preference differences between caption pairs are substantially smaller.

Our results highlight a broader challenge in AI capacity to understand subgroup and individual preferences for subjective and creative tasks. In Section~\ref{sec:position}, we argue that the AI research community's focus on problems with verifiable rewards, in domains such as mathematics and coding, may be insufficient for achieving AGI. We propose that mastering creative domains -- which lack objective metrics and require deep understanding of audience preferences -- represents a crucial yet underexplored challenge on the path to AGI. 

The contributions of this work are as follows.
\begin{enumerate}
    \item We decompose LLM capability in humorous caption ranking into three fundamental components -- visual understanding, humor reasoning and subgroup preference alignment.
    \item By improving upon all of the three components, especially on the preference alignment, we obtain caption ranking models that achieve accuracy on par with human experts.
    \item Our experiments reveal that extensive persona-based alignment significantly under-performs relative to improvements based on finetuning, revealing current LLM limitations in understanding subgroup and individual preferences.
    \item We propose that systematic collection and integration of human preference data across creative domains may be essential for achieving AGI in creative tasks.
\end{enumerate}

\section{Related Work}
\begin{figure*}[ht]
    \begin{minipage}[b]{0.5\textwidth}
        \centering
        \includegraphics[width=\linewidth]{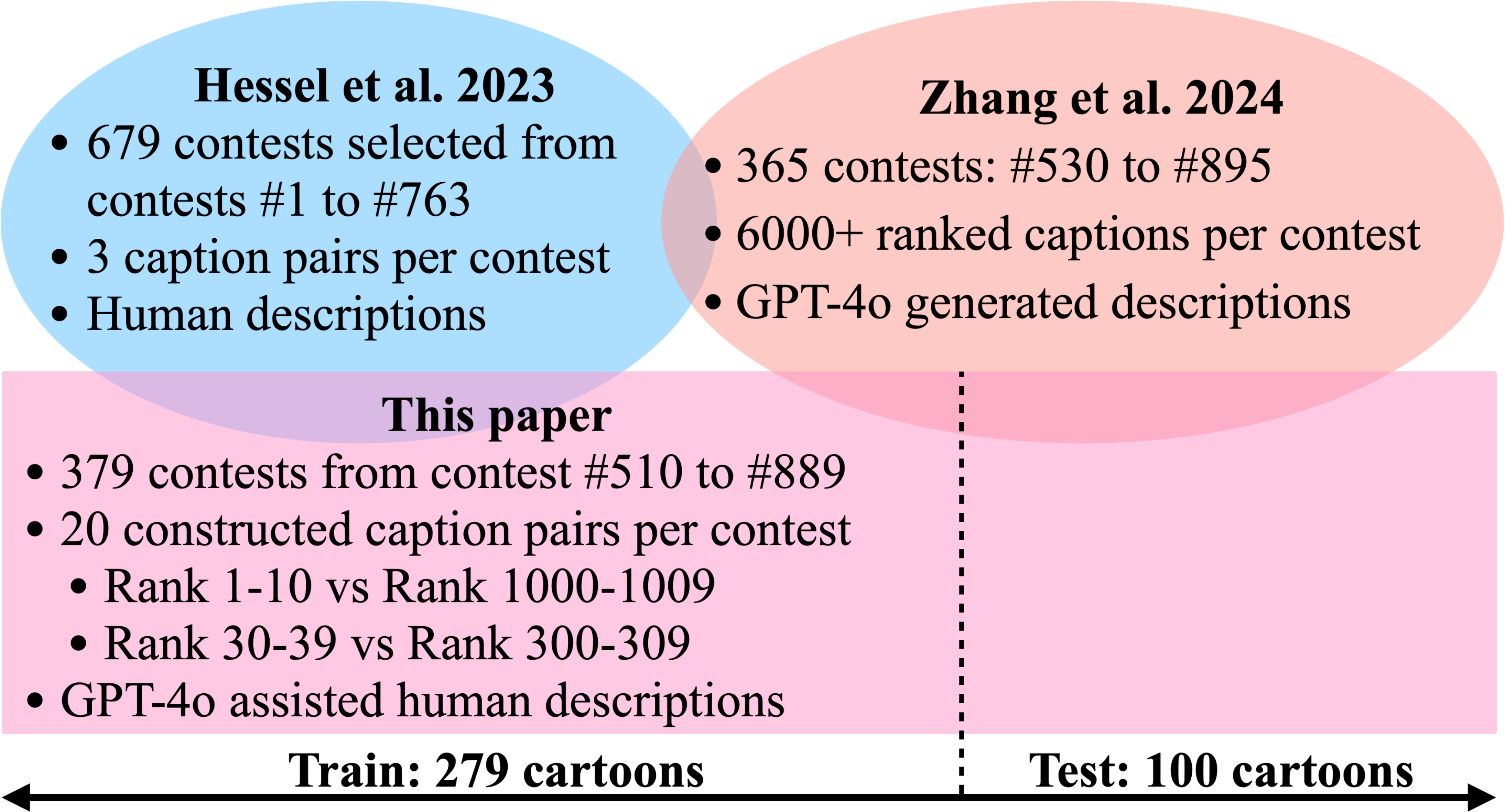}
        \caption{Composition of cartoon caption contest datasets across \citet{hessel-etal-2023-androids}, \citet{zhang2024humor} and our paper. In our paper, we examine $20$ pairs of captions selected from 379 contests (\#510-\#889). The dataset is further split into 279 contest for training and 100 for testing.}
        \label{fig:dataset visualization}
    \end{minipage}
    \hfill
    \begin{minipage}[b]{0.47\textwidth}
        \centering
        \includegraphics[width=\linewidth,trim={3cm, 2cm, 3cm, 1cm},clip]{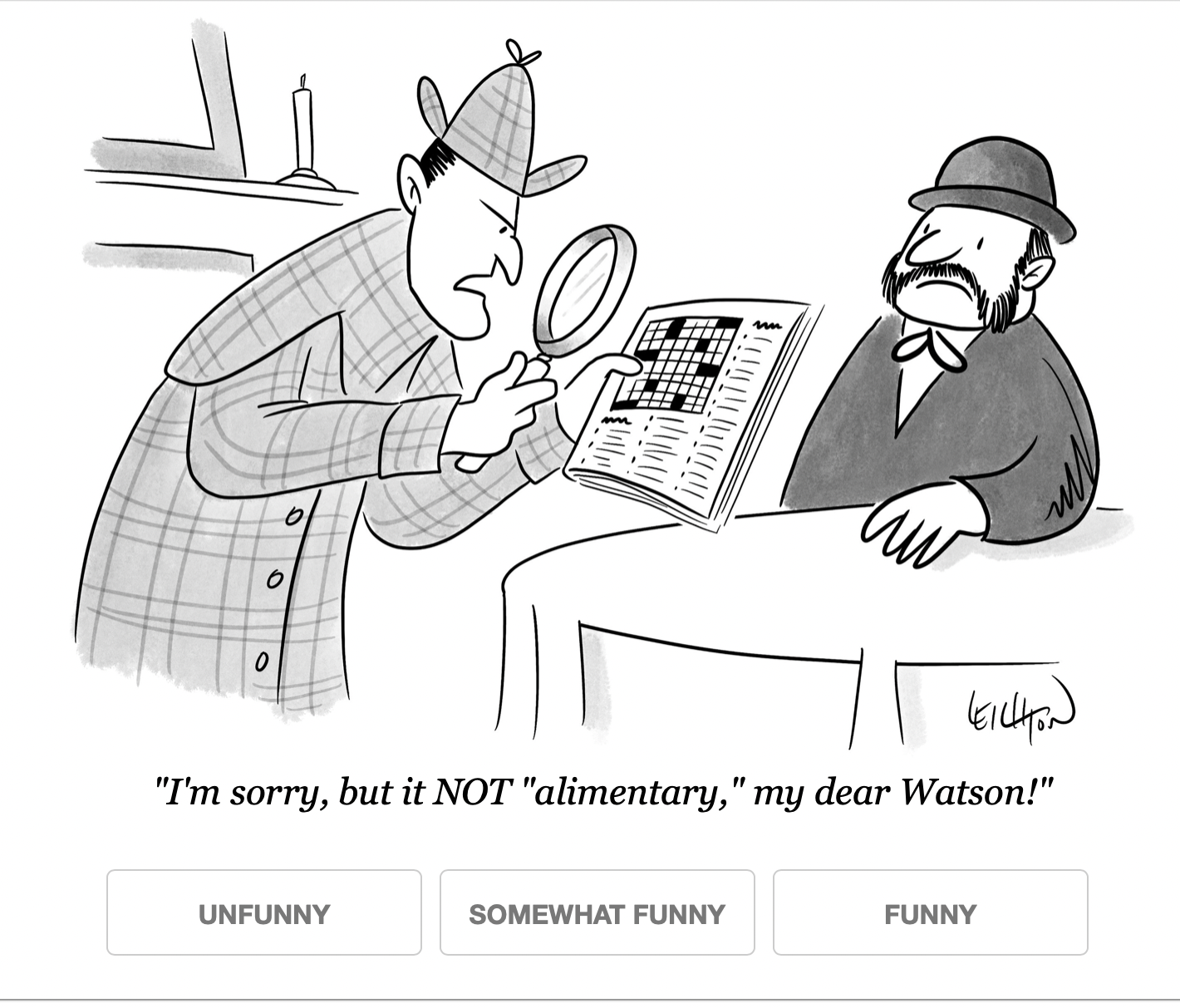}
        \caption{Example voting page for the caption contest.}
    \label{fig:voting}
    \end{minipage}
    \vspace{-\intextsep}
\end{figure*}

\textbf{Humor and LLMs.}
Research on computational humor has evolved significantly -- from early rule‐based, template‐driven systems that generated puns via fixed linguistic rules \cite{Binsted2006, Apte1988} to modern large language models (LLMs) that strive to capture the nuances of human wit. Recent studies reveal that while models like ChatGPT can produce coherent and seemingly humorous outputs, they often rely on a limited repertoire of pre‐learned jokes rather than inventing truly original humor \cite{JentzschKersting2023}. To overcome these limitations, innovative prompting strategies such as the Creative Leap-of-Thought (CLoT) paradigm have been proposed, encouraging LLMs to make unexpected conceptual associations and thereby enhancing creative humor generation \cite{Zhong2023}. Complementing these approaches, multimodal techniques that integrate auditory cues have shown promise in capturing the phonetic ambiguities (essential for understanding puns) that text-only systems often miss \cite{Baluja2025}. Furthermore, research on curated humor datasets demonstrates how targeted data can expose LLM limitations and spur advances in humor generation \cite{Horvitz2024}, while real-world evaluations by stand-up comedians underscore that, despite impressive fluency, LLM outputs frequently appear generic or bland compared to human creativity \cite{Mirowski2024}. On the other hand, recent studies suggest that under controlled conditions AI-generated humor can rival human-produced jokes \cite{NYPost2024}. Challenges remain, however, in producing humor that is contextually rich, culturally sensitive, and genuinely surprising, highlighting the need for continued research into more sophisticated models and training paradigms \cite{LastLaugh2024}.

\textbf{New Yorker Cartoon Caption Contest.}
Recent advances in computational humor have been bolstered by the availability of large, well‐curated datasets derived from The New Yorker Cartoon Caption Contest. Previous works used this dataset to analyze the complex interplay between visual cues and linguistic humor, shedding light on the mechanisms that make captions amusing \cite{zhang2024humor}. The seminal work of Bob Mankoff, whose editorial work shaped the contest’s creative process, provides essential context and insight into what constitutes successful humor in this setting \cite{Mankoff2008}. However, recent studies have demonstrated that state-of-the-art AI models struggle to fully capture the nuanced judgment required to select and explain winning captions \cite{hessel-etal-2023-androids}. Together, these works underscore the utility of the New Yorker dataset as a powerful benchmark for advancing our understanding of humor in both human and machine-generated contexts.

\textbf{LLM Post-training/Alignment}
Recent advancements in post-training alignment techniques for LLMs have progressed through several distinct stages. Initially, supervised fine-tuning (SFT) was employed to adapt pre-trained models to specific tasks using high-quality, instruction-based datasets, demonstrating that even modest amounts of curated data can substantially improve downstream performance \cite{wei2021finetuned}. Building on this, researchers introduced Reinforcement Learning from Human Feedback (RLHF) \citep{ouyang2022training}to further align model outputs with human preferences. In this framework, Proximal Policy Optimization (PPO) is widely used to adjust the model’s behavior based on human-provided preference comparisons \cite{schulman2017proximal}. However, the inherent complexity and instability of PPO-based RLHF motivated the development of simpler alternatives. Direct Preference Optimization (DPO) recasts the alignment objective as a supervised learning problem by directly contrasting the log-probabilities of preferred and non-preferred responses, thereby eliminating the need for an explicit reward model \cite{rafailov2023direct}. More recently, extensions such as Group Relative Policy Optimization (GRPO) have been proposed, which incorporate group-level comparisons that further enhance training stability and mitigate issues like catastrophic forgetting \cite{guo2024grpo}. This evolution -- from SFT through PPO-based RLHF to DPO and GRPO -- reflects the field’s ongoing efforts to develop robust, efficient, and reliable post-training alignment methods for LLMs.

\section{Cartoon Caption Ranking Task}

The New Yorker Cartoon Caption Contest is a long-standing weekly feature hosted by The New Yorker magazine, in which a captionless cartoon is published and readers are invited to submit humorous captions. Each week, over 6,000 captions are submitted. From contest \#530 to contest \#895, a bandit-based crowdsource rating system~\citep{jamieson2015next} has been employed, allowing users to score captions as ``funny'', ``somewhat funny", or ``unfunny'' (see Figure~\ref{fig:voting}). At the end of each contest, a complete crowdsourced ranking of captions is obtained based on their perceived humor. Over the past eight years, the dataset of cartoons, captions and their rankings~\citep{zhang2024humor} has proven invaluable for computational humor research. Notably, prior work by \citet{hessel-etal-2023-androids} and \citet{zhang2024humor} has leveraged the caption contest dataset to benchmark both humor understanding and generation, two essential domains of humor reasoning.

To evaluate caption understanding, we employ the pairwise ranking task, a method widely used to study humor \citep{shahaf2015inside,radev2015humor,king2013random,hessel-etal-2023-androids,zhang2024humor}. We adopt the variant described by \citet{hessel-etal-2023-androids}. In this task, given a cartoon description\footnote{A cartoon image was used in place of the cartoon description when humor experts performed the same task.}, evaluators (models or humans) compare two captions at a time, each randomly sampled from distinct ranking tiers. Specifically, one caption is drawn from a high-ranked group (ranks \#1–10) and the other from a lower-ranked group (ranks \#1000–1009) (see Figure~\ref{fig:dataset visualization}). This sampling strategy allows us to directly measure an evaluator's ability to discern differences in humor quality while controlling task difficulty through the selection of ranking tiers. Additionally, we conduct a more challenging variant by asking models to distinguish between captions sampled from mid-ranked tiers (ranks \#30–39 versus ranks \#300–309). Previous work by \citet{hessel-etal-2023-androids} and \citet{zhang2024humor} shows that state-of-the-art models, including variants of GPT-4, achieve only around 67\% accuracy on the easy version of the pairwise task, whereas human experts significantly outperform the LLMs. These findings underscore the persistent gap between current state-of-the-art models and human expertise in humor understanding, motivating our investigation into novel approaches to enhance model performance on this task.

\section{Experiments} 

We break the ranking challenge into three components -- visual understanding, humor reasoning, and targeted alignment to human crowd preferences.

Generating cartoon description is a crucial first step in understanding the humor correctly. However, we found 23.5\% of the GPT-4o generated cartoon descriptions in \citet{zhang2024humor} have erroneous descriptions. Therefore, in Section~\ref{ssec:annatation}, we employ an AI-assisted annotation with human-in-the-loop assistance to fix cartoon descriptions.
In Section~\ref{ssec:explanation}, we find that the o1-preview model can explain captions correctly and demonstrates extensive humor reasoning more than 85\% of the time. We therefore generate such explanations, which serve as intermediate reasoning steps that inform the final pairwise comparison of captions.
To better align our system with human crowd preferences, we implement two different strategies in Section~\ref{ssec:alignment}. First, we conducted extensive persona-based system prompting, which does not exhibit significant improvements. Our second, more sophisticated approach directly fine-tunes the model based on a set of ground truth rankings collected in the crowdsource ranking. This second approach significantly improves the ranking accuracy, and closes the performance gap between LLMs and human experts.

Throughout this section, we use a random train/test split (see Figure~\ref{fig:dataset visualization}) with 279 cartoons for training and 100 for testing. The training set is also used for sampling 5-shot in-context learning, with five meaningfully sampled caption pairs per cartoon. All reported results are evaluated on the test set.

\begin{figure*}[ht]
\centering
\includegraphics[width=1\textwidth, trim={0 0.6cm 0 0.6cm},clip]{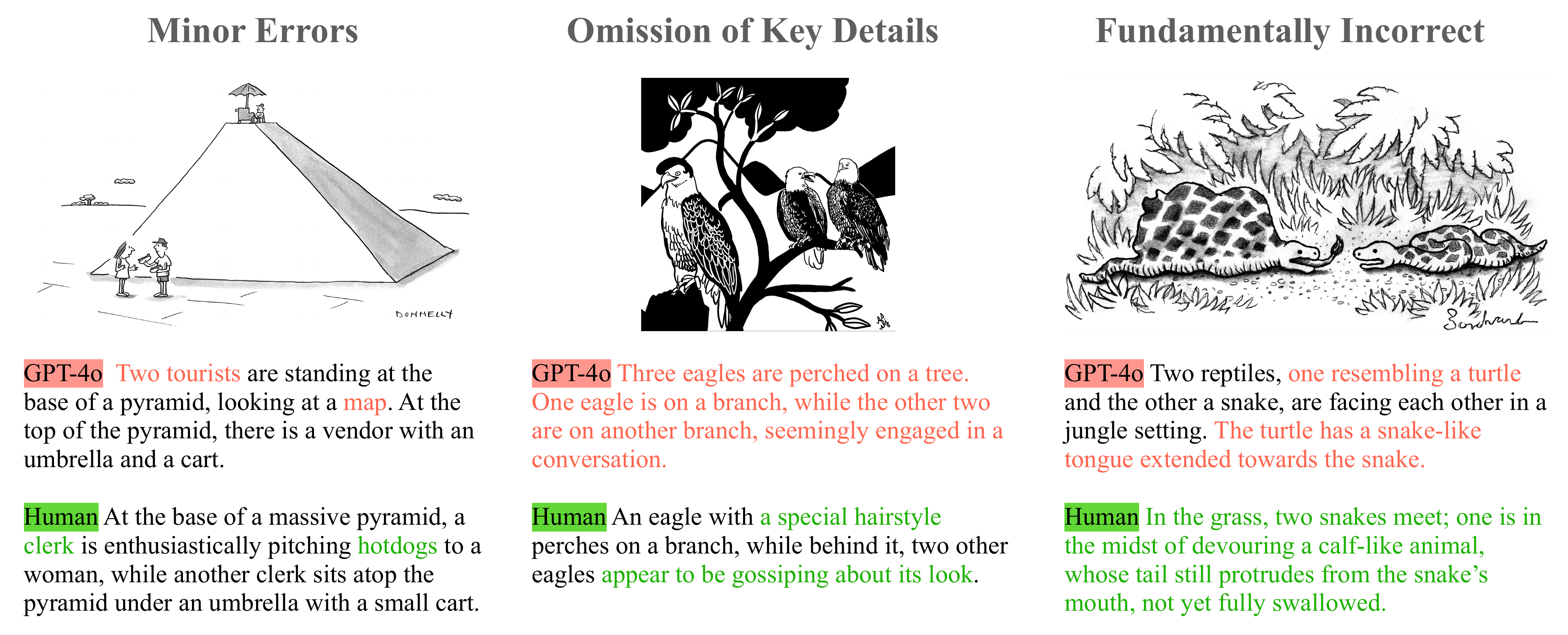}
\caption{Examples of three types of errors in machine-generated cartoon descriptions and their human-annotated corrections. Left: Minor errors in word choice ("tourists" vs. "clerk", "map" vs. "hotdogs"). Center: Omission of key narrative details (missing the humorous implication of eagles gossiping about another eagle's appearance). Right: Fundamentally incorrect scene interpretation (misidentifying two snakes as a turtle and snake).}

\vspace{-10pt}
\label{fig:description}
\end{figure*}
\subsection{Improved Visual Annotation} \label{ssec:annatation}
Our dataset comprises 379 cartoons from the caption contest, including a subset from the annotated dataset introduced by \citet{hessel-etal-2023-androids}. For cartoons lacking human annotations, we extend the description generation approach of \citet{zhang2024humor}. Through an LLM-assisted annotation process, we refine and improve the existing cartoon descriptions to build a comprehensive dataset.

Our visual annotation aims to generate both canny and uncanny descriptions. The canny descriptions accurately capture the literal contents of a cartoon, while the uncanny descriptions highlight its unusual or unexpected elements.

Our quality assessment reveals that 23.5\% (89/379) of the machine-generated descriptions contain inaccuracies of varying severity, ranging from minor semantic errors and missing contextual elements to fundamental misinterpretations of the scene (see Figure~\ref{fig:description}). To address these issues, we develop a two-phase annotation refinement process. In the first phase, human reviewers iteratively improve the canny descriptions by identifying and correcting incorrect or omitted details. Based on their feedback, the descriptions undergo targeted revisions until they achieve comprehensive accuracy. In the second phase, these validated canny descriptions are used to generate corresponding uncanny elements, ensuring analytical consistency throughout the annotation process. Further details on this process can be found in Appendix~\ref{apx:update_description}.

Comparative experiments between using the original and refined descriptions show an accuracy improvement from 70\% to 73\% with GPT-4o prompting. With the refined descriptions, finetuned models (more details in Section~\ref{sec:finetuning section}) obtain a performance gain from $81.3\%$ to $82.4\%$.

\subsection{Does reasoning through a joke improve humor understanding?} \label{ssec:explanation}

Humor assessment is challenging because it fuses objective cues with subjective preferences. While enhanced caption descriptions capture objective elements, they often miss the figurative aspects that make a caption truly funny.

We propose that enriching model inputs with explicit explanations can improve performance by highlighting both objective cues (e.g., wordplay) and subjective nuances (e.g., cultural context). Prior work \citep{hessel-etal-2023-androids} shows that models detect objective features well but struggle with subjectivity. Encouragingly, recent reasoning models like o1 and DeepSeek \cite{openai2024o1, deepseek2024} appear promising -- our humor expert found that over 85\% of o1-preview explanations effectively captured a cartoon’s humor.



\begin{figure*}[t]
\centering
\includegraphics[width=1\textwidth]{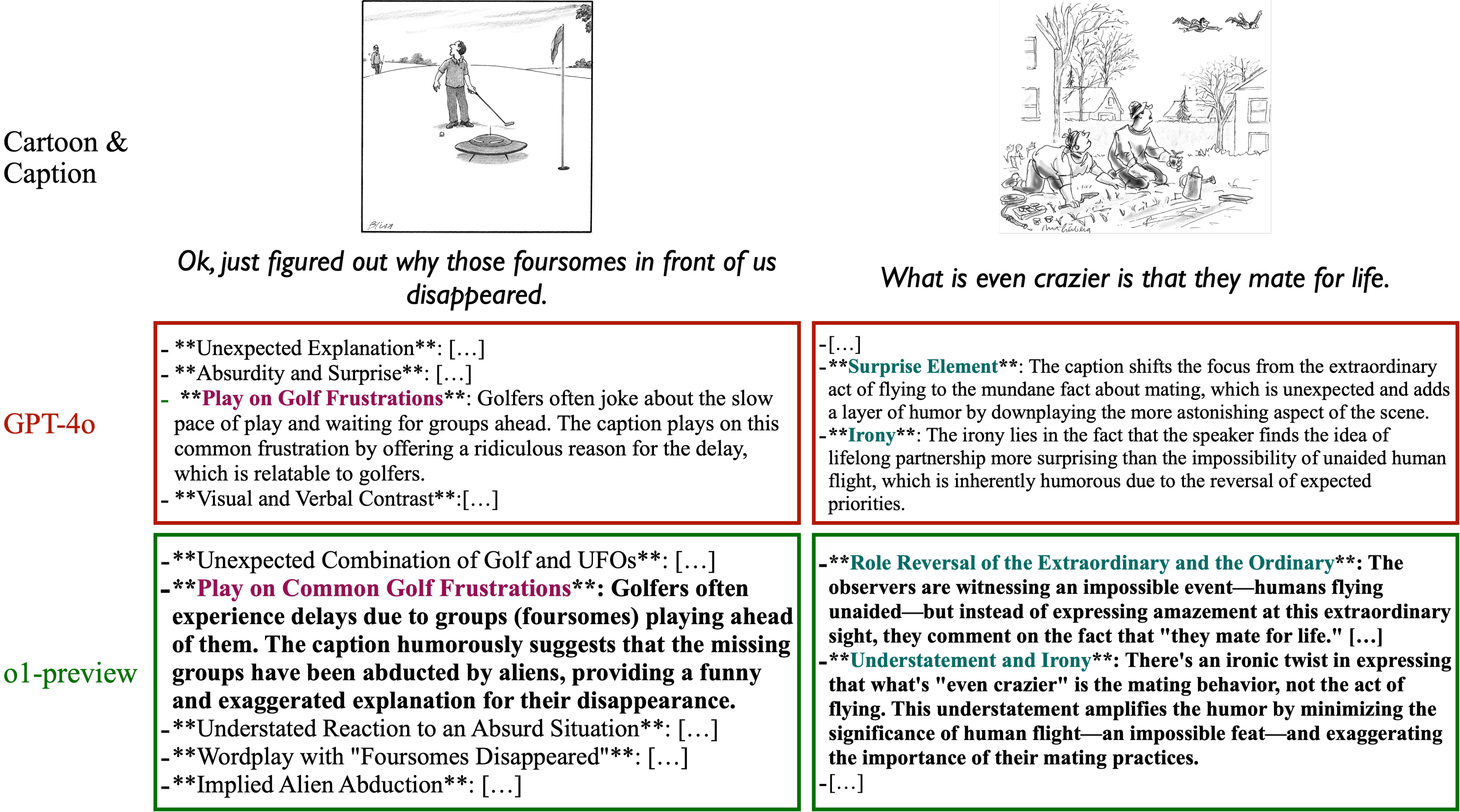}
\caption{Comparison of humor explanation quality between GPT-4o and o1-preview, illustrated through two cartoon-caption pairs and their respective AI-generated humor explanation. o1-preview demonstrates a deeper comprehension of the humor, and its explanations are highlighted in bold text.
}
\label{fig:explanations}
\vspace{-\intextsep}
\end{figure*}

We generate explanations using two language models, GPT-4o and o1-preview (see Appendix~\ref{apx:explanation_generation}). As shown in Table \ref{tab:reasoning}, o1-preview explanations boost ranking accuracy to 76\%, compared to 73\% for the baseline and 71\% for GPT-4o-generated explanations. This underscores the importance of explanation quality, with o1-preview better capturing humor nuances (see Figure \ref{fig:explanations}).

\begin{table}[t]
    \begin{center}
        \begin{tabular}{lc}
        \toprule
     Explanation Model & Accuracy \\
    \hline
    none (baseline) & 73\%\\
     GPT-4o   & 71\%\\
     o1-preview & \textbf{76\%} \\
    \bottomrule
    \end{tabular}
    \end{center}
    \caption{GPT-4o pairwise caption ranking accuracy of top 10 vs 1000-1009 captions. We compare explanations generated by different models. The experiment is conducted with five in-context examples (detailed prompts in Appendix~\ref{apx:explanation_icl_prompts}).}
    \label{tab:reasoning}
    \vspace{-\intextsep}
\end{table}

Our findings indicate that equipping the ranking model with explicit explanations bridges the gap between objective cues and subjective humor elements. However, accuracy still falls short of expert human performance (90\%, Table~\ref{tab:combined_accuracy}), possibly due to the unique subjective biases of the Caption Contest ranking crowd. In the next section, we explore aligning the verifier's preference with that of the New Yorker Caption Contest crowd to further narrow this gap.

\subsection{Alignment to Crowd Preference} \label{ssec:alignment}
Despite our success in incorporating both objective structural cues and subjective reasoning-based explanations in previous sections, a significant performance gap remains between our models and humor experts. Our experiments in the pairwise caption ranking task demonstrate that while adding these elements enhances humor understanding, the verifier model (GPT-4o) still does not fully capture the nuanced humor preferences of the New Yorker caption contest crowd. We hypothesize that this discrepancy arises from a fundamental misalignment: GPT-4o’s inherent subjective reasoning does not match the specific taste and evaluative criteria of New Yorker voters.

To address this misalignment, we build on our previous findings (that integrating objective and subjective elements enhances humor understanding) by exploring two alignment strategies. The first approach employs persona-based prompting to simulate the subjective evaluative criteria of active New Yorker caption contest participants, subtly steering the model’s preference toward the target audience’s distinctive preferences. Our second strategy takes a more direct route through supervised fine-tuning (SFT) on a large corpus of New Yorker caption contest caption ranking data released by \citet{zhang2024humor}. We hypothesize that direct adaptation to the target domain will better align the model’s humor judgments with those of New Yorker voters, thereby narrowing the performance gap with human experts.

In the following sections, we detail these two alignment strategies and evaluate their effectiveness in bridging the gap between our model’s performance and the nuanced humor understanding of human experts.

\subsubsection{Persona-Based Prompting} \label{sec:persona section}
Persona-based prompting embeds persona information within system prompts to steer language model outputs toward reflecting target audience preferences, emerging as a promising method for aligning model behavior. Prior research has demonstrated the effectiveness of this approach in various tasks \citep{park2023generative, chuang2023simulating, chen2024persona, chuang2024wisdom}. We design nine distinct prompts that simulate personas representative of the New Yorker Cartoon Contest audience and evaluate their impact on humor preference alignment using four language models—Claude-3.5-sonnet, Gemini, GPT-4o, and o3-mini \cite{anthropic2024claude, google2024gemini, openai2024gpt4o, openai2025o3mini}.

\begin{table*}[t]
    \centering
    \begin{tabular}{ccccc}
        \toprule
        \textbf{System Prompt} & \textbf{GPT-4o (\%)} & \textbf{Claude (\%)} & \textbf{Gemini (\%)} & \textbf{o3-mini (\%)} \\
        \midrule
        Empty (Baseline)  & 73.5 & 68 & 47 & 70 \\ 
        Judge & 73 & 67.5 & 55 & 58 \\ 
        Male Lawyer & 75 & 74 & 51 & 59 \\
        Female Lawyer & \textbf{76.5} & 69 & 59 & 67 \\
        CS Phd & 73.5 & 68 & 49 & 65 \\
        Sociologist \& Psychologist & 73.5 & 67 & 61 & 60 \\
        Literature Student & 73.5 & 72 & 51 & 58 \\
        Bob Mankoff & 73.5 & 66 & 50 & 62 \\
        Larry Wood & 73 & 68 & 57 & 61 \\
        Cartoon Author & 71.5 & 61 & 46 & 62 \\
        \bottomrule
    \end{tabular}
    \caption{Performance of using different persona-based system prompts on 10 vs 1000 pairwise caption ranking task across four language models: GPT-4o, Claude-3.5-Sonnet, Gemini-2.0-Thinking-Experiment, and o3-mini. Each number is measured on a size $200$ subset of the test set. Each row represents a distinct persona-based prompt. See Appendix~\ref{apx:system_prompt_details} for system prompts and Appendix~\ref{apx:explanation_icl_prompts} for the task prompt.}
    \label{tab:system_prompts}
    \vspace{-\intextsep}
\end{table*}

In all of our persona-based experiments, we incorporate five random in-context learning examples, and also the o1-preview explanations. See Appendix~\ref{apx:explanation_icl_prompts} for detailed prompts. Results in Table \ref{tab:combined_accuracy} show that persona-based prompts yield only modest improvements in aligning the model with the intended audience, with the highest accuracy of 76.5\% achieved using the Female Lawyer prompt—a mere 3\% gain over the baseline without any persona. These results indicate that persona-based alignment is not strong enough to capture the preferences of the New Yorker crowd. Instead, more powerful alignment strategies, such as supervised fine-tuning (SFT), are required. In the next section, we detail our application of SFT to better align the model with human preferences and bridge the performance gap.

\begin{table}[t]
    \begin{center}
        {
        \setlength{\tabcolsep}{2pt} 
        \scalebox{.95}{
        \begin{tabular}{lcc}
            \toprule
            Methods & 10vs1000 & 30vs300 \\ 
            \midrule
            Expert Majority Vote  & $84_{\pm5.2}$   & $66_{\pm6.8}$   \\
            Expert Average Accuracy & $78_{\pm2.6}$   & $61.6_{\pm3.0}$   \\ 
            Best Expert Accuracy  & $85.33_{\pm2.9}$   & $68_{\pm3.8}$   \\ \midrule
            Gemini 2.0 Flash Thinking & $51.8_{\pm1.6}$ & $50.6_{\pm1.6}$\\
            o1 & $69_{\pm1.5}$ & $58_{\pm1.6}$\\
            o3-mini & $60.7_{\pm1.5}$ & $53.2_{\pm1.6}$\\
            Claude 3.5 Sonnet & $65.8_{\pm1.5}$ & $54_{\pm1.6}$\\
            GPT-4o Prompting & $67.3_{\pm1.5}$ & $53.9_{\pm1.6}$\\
            GPT-4o SFT w/o Expl. (Ours) & $79.4_{\pm1.3}$ & $59.7_{\pm1.6}$\\
            GPT-4o SFT w/ Expl. (Ours) & $\mathbf{82.4_{\pm1.2}}$ & $\mathbf{63.2_{\pm1.5}}$\\
            \bottomrule
        \end{tabular}
        }}
        \caption{Accuracy(\%) Comparison of Different Methods. All models (Gemini, o1, o3-mini, Claude and GPT-4o) use prompting techniques including best persona, 5-shot in-context learning, and o1-preview generated explanations. Our fine-tuned GPT-4o model with humor explanations even outperforms the best human expert.}
        \label{tab:combined_accuracy}
    \end{center}
    \vspace{-2\intextsep}
\end{table}

\subsubsection{GPT-4o Finetuning} \label{sec:finetuning section}
Despite the limited accuracy gain by persona-based prompting, aligning to a specific group of audience should not be a hard task especially when the model is given access to the correct understanding and explanations of the captions. Indeed, once we finetune SOTA LLMs on a small set of human preferences, we recover a significantly higher ranking accuracy. This also reveals an interesting failure case where LLMs fail in understanding subgroup and individual human preferences for subjective tasks (more discussions in Section~\ref{sec:position}).

To construct the training set, for each of the 279 training cartoons, we randomly form 10 pairwise comparisons between captions ranked 1-10 and those ranked 1000-1009, and another ten pairs between 30-39 and 300-309. This in total results in 5580 pairs of captions. In our experiments, we performed a simple supervised finetuning of GPT-4o, for it to choose between the two candidate captions. The model is given the cartoon description, the two captions and their corresponding explanations generated by o1-preview. 

Note that pairwise comparisons of captions ranked 30-39 versus those ranked 300-309 have a much narrower gap, and are thus much more challenging. To the best of our knowledge, we are the first to evaluate performance on this task. Results on both the easier (1-10 vs 1000-1009) and the the more challenging (30-39 vs 300-309) sets of comparisons are reported. As shown in Table~\ref{tab:combined_accuracy}, fine-tuned GPT-4o models can significantly improve upon all prompting-based baseline before. When incorporating o1-preview generated explanations, the finetuned GPT-4o can achieve slightly higher accuracy than the average over human experts. Below, we give some details on the human expert experiments.

\subsubsection{Human Expert Accuracy}
To evaluate the performance of our model, we conducted a study with five highly renowned human experts in the New Yorker Cartoon Caption Contest world, including famous cartoonists, editors and podcast hosts in this area. In our experiments, these human experts were presented with both the original cartoon image and paired captions, tasked with selecting the more humorous option from each pair (demonstrated in Figure~\ref{fig:schematic}). Due to busy schedules of these experts, the evaluation corpus consisted of 50 contests selected from our testing set, with two distinct caption pairs each. The two pairs are consisted of one for comparison between the top-ranked caption (rank 1) and a lower-ranked caption (rank 1000), and a comparison between a mid-ranked caption (rank 30) and a lower-ranked caption (rank 300). 

The result in Table~\ref{tab:combined_accuracy} shows the accuracy of the majority vote among the five experts as well as the average of their individual performances. In addition, the best individual performance was from Bob Mankoff, the former chief cartoon editor for the New Yorker, who created this contest more than $25$ years ago. Our model still slightly underperforms the performance Mankoff, leaving space for further improvement. More details about our expert experiments can be found in Appendix~\ref{apx:human_stats}, where we do see strong inter-rater agreement among human experts.

\section{Position: To achieve AGI, LLMs require much more human interaction data to acquire the understanding of individual/subgroup level preferences.} \label{sec:position}

Our empirical findings on humor comprehension point to a broader challenge in artificial intelligence: the development of true creative understanding. While recent advances in LLMs have demonstrated remarkable capabilities in analytical reasoning and structured problem-solving, our results suggest that creative domains may present a unique and potentially final hurdle in achieving AGI. We argue that this challenge stems from two fundamental characteristics of creative tasks that are often overlooked in current AI research.

First, creative tasks inherently lack verifiable rewards. Unlike mathematical proofs or programming challenges where correctness can be definitively verified, creative success often depends on subjective human judgment. Our experiments with the New Yorker Caption Contest illustrate this clearly: while our models can now generate sophisticated explanations of why a caption might be humorous, these explanations alone do not translate to accurate predictions of human preferences. This suggests that current approaches to AI alignment, which often focus on optimizing for verifiable metrics, may be insufficient for creative domains.

Second, and perhaps most challengingly, creative excellence requires understanding and internalizing group-specific preferences and cultural contexts. Our finding that persona-based prompting failed to improve caption ranking, while direct preference learning proved effective, highlights a crucial gap in current LLM capabilities. While the New Yorker Caption Contest provides us with extensive ranking data from a specific audience, collecting similarly comprehensive preference data for every creative domain, cultural group, and individual taste remains prohibitively difficult. For instance, how might we gather equivalent preference data for domains like musical composition, architectural design, or scientific research, where expert judgment is highly specialized and preferences can vary dramatically across different communities?

These observations lead us to propose that achieving AGI may fundamentally require solving the challenge of preference understanding. While we can use reinforcement learning and inference-time scaling techniques to improve creative generation once we have reliable judgment models, the path to AGI requires models that can develop generalizable insights about how preferences function across different contexts and domains. This suggests that just as human creativity is deeply influenced by understanding others' perspectives and preferences, AGI systems will need to develop a fundamental grasp of how preferences work—not just in individual domains, but as a generalizable concept—to achieve truly intelligent creative behavior.

\section{Conclusion}
Our work demonstrates that by decomposing humor understanding into visual comprehension, reasoning, and preference alignment components, LLMs can achieve expert-level performance in humor evaluation. While persona-based prompting showed limited success, direct fine-tuning on crowd preferences yielded dramatic improvements, suggesting that systematic collection of human preference data across creative domains may be essential for achieving true creative understanding in AI systems. Looking ahead, our high-performing model can serve as a reliable verifier for humor generation, enabling inference-time scaling techniques \citep{Zelikman2022STaR,snell2024scaling} to improve creative output. This creates a promising pathway for advancing both humor understanding and generation capabilities in AI systems.

\section{Limitations}

Our work has several limitations that we acknowledge below:
\begin{itemize}
    \item \textbf{Domain Specificity.} Our study is based solely on the New Yorker Cartoon Caption Contest dataset. Although this dataset provides a rich benchmark for humor evaluation, its narrow focus may limit the generalizability of our findings to other forms of humor and creative tasks.
    
    \item \textbf{Evaluation Focus.} We primarily evaluate caption understanding using a pairwise ranking task. While this approach is effective for assessing relative humor quality, it may not fully capture the broader nuances of humor understanding or the challenges involved in humor generation.
    
    \item \textbf{Subjectivity and Bias in Preference Data.} The human preference data employed for fine-tuning and evaluation is inherently subjective and reflects the tastes of a specific audience (e.g., New Yorker readers). This limitation, however, reinforces our position that systematic collection of diverse human preference data is crucial for improving model performance on creative tasks.
    
    \item \textbf{Scalability of Human Alignment.} While our results demonstrate that aligning models with human preferences can substantially enhance performance, the process of gathering high-quality, curated human data is resource-intensive and may not scale easily to other creative domains. This challenge underlines our broader argument that advancing creative AI requires scalable methods for collecting and integrating human interaction data.

    \item \textbf{Humorous Content May Be Offensive.} Humor often walks a fine line between eliciting laughter and being potentially offensive. While our focus on the New Yorker dataset biases our work towards a certain style of humor, we acknowledge that humorous content can sometimes be culturally insensitive or derogatory. Our current framework does not explicitly address the detection or mitigation of offensive content, highlighting the need for future research to incorporate robust ethical safeguards alongside creative performance.
\end{itemize}


\section*{Acknowledgments}
We would like to thank Vincent Coca, Beth Lawler, Joel Mishon and Paul Nesja for their expert annotation effort (in addition to Bob Mankoff's), which establishes a strong baseline of how well human can do on this task.

\clearpage
\bibliography{custom}

\clearpage
\appendix
\onecolumn


\section{Language Model Prompt}\label{apx:prompt}

\subsection{Updating description}\label{apx:update_description}
We conduct a comprehensive quality assessment of the cartoon descriptions generated by GPT-4o across our dataset of 379 images. Initial evaluation reveals that 76.5\% of the generated descriptions meet our quality criteria for reasonableness and completeness. The remaining 23.5\% exhibits various types of deficiencies that required remediation.

To address these quality issues, we implement a systematic two-phase refinement process: 

In the first phase, for the identified problematic descriptions, we provide GPT-4o with specific feedback detailing the observed errors and request regeneration of these descriptions. This iterative process continues until the descriptions achieve the required level of accuracy and completeness.

In the second phase, following the establishment of a clean description set, we employ a 5-shot learning approach to generate corresponding uncanny descriptions for those updated canny descriptions. The following is a detailed prompt of the second phase. 
\begin{center}
\fbox{
\begin{minipage}{.95\linewidth}
\textbf{User}: In this task, you will see a cartoon image and a canny description written about the image. You need to write an uncanny description. I’m going to give you five examples first. Write an uncanny description for the last set.

\textbf{User}: $<$Insert Cartoon Image$>$ 

\textbf{User}: The canny description is $<$Insert canny description$>$

\textbf{Assistant}: The uncanny description is $<$Insert uncanny description$>$

......\textit{Repeat user/assistant for four more examples}......

\textbf{User}: $<$Insert Cartoon Image$>$ 

\textbf{User}: The canny description is $<$Insert canny description$>$

\textbf{User}: The uncanny description is

\end{minipage}
}
\end{center}

\subsection{Explanation Generation}\label{apx:explanation_generation}
We employ both GPT-4o and o1-preview to generate explanations for the humorous elements in the captions. We implement a zero-shot approach, providing each model with both the uncanny and cannny descriptions alongside the caption in question. The prompt structure utilized in our experiments is illustrated below.

\begin{center}
\fbox{
\begin{minipage}{.95\linewidth}
\textbf{User}: I will give you a description of the cartoon and the winning caption. Explain to me why the caption is funny.

\textbf{User}: The descriptions for the images are $<$Insert canny description$>$ and $<$Insert uncanny description$>$ The winning captions is: $<$Insert cartoon captions$>$

\textbf{User}: There may or may not be multiple reasons for the caption being funny. Put them into bullet point(s).
\end{minipage}
}
\end{center}

\subsection{Baseline Caption Evaluation}\label{apx:baseline_prompts}
For our baseline evaluation, we employ a 5-shot prompting approach. In this setup, we provide the model with cartoon descriptions and the corresponding pair of captions. The prompt follows a structured format where the model is first assigned the role of a judge for the New Yorker cartoon caption contest. We then present five examples of caption ranking, allowing the model to observe the evaluation pattern. For the final test case, the model is tasked with selecting the funnier caption between two options, as the examples. The prompt structure is illustrated below.

\begin{center}
\fbox{
\begin{minipage}{.95\linewidth}
\textbf{System}: You are a judge for the New Yorker cartoon caption contest.

\textbf{User}: In this task, you will see two descriptions for a cartoon. Then, you will see two captions that were written about the cartoon. Then you will choose which caption is funnier. I am going to give you five examples first and you answer the last question with either A or B

\textbf{User}: For example, the descriptions for the images are $<$Insert canny description$>$ and $<$Insert uncanny description$>$. The two captions are A: $<$Insert Caption A$>$. B: $<$Insert Caption B$>$

\textbf{Assistant}: The caption that is funnier is $<$Insert Answer$>$

......\textit{Repeat user/assistant for four more examples}......

\textbf{User}: The descriptions for the images are $<$Insert canny description$>$ and $<$Insert uncanny description$>$. The two captions are A: $<$Insert Caption A$>$. B: $<$Insert Caption B$>$

\textbf{User}: Choose the caption that is funnier. Answer with either A or B and nothing else.
\end{minipage}
}
\end{center}

\subsection{ICL Explanation Caption Evaluation}\label{apx:explanation_icl_prompts}
Building on the baseline evaluation, we incorperate o1-preview generated  the model an explanation that is generated by o1. We changed the system prompts to test different persona. The detailed persona prompts is in Appendix~\ref{apx:system_prompt_details}.

\begin{center}
\fbox{
\begin{minipage}{.95\linewidth}
\textbf{System}: You are a judge for the new yorker cartoon caption contest.

\textbf{User}: In this task, you will see two descriptions for a cartoon. Then, you will see two captions about the cartoon and an explanation for why each caption is funny. I am going to first give you five examples where I will tell you which one is funnier then you answer the last one with either A or B and nothing else.

\textbf{User}: For example, the descriptions for the images are $<$Insert canny description$>$ and $<$Insert uncanny description$>$. Captions A: $<$Insert Caption A$>$, and why the caption is funny is $<$Insert explanation for Caption A $>$. Caption B: $<$Insert Caption B$>$, and why the caption is funny is $<$Insert explanation for Caption B$>$,

\textbf{Assistant}: The caption that is funnier is $<$Insert Answer$>$

......\textit{Repeat user/assistant for four more examples}......

\textbf{User}: Last one, the descriptions for the images are $<$Insert canny description$>$ and $<$Insert uncanny description$>$. Caption A: $<$Insert caption A$>$, and why the caption is funny is $<$Insert explanation for Caption A$>$. Caption B: $<$Insert Caption B$>$, and why the caption is funny is $<$Insert explanation for Caption B$>$.

\textbf{User}: The caption that is funnier is 
\end{minipage}
}
\end{center}
\begin{table*}[ht!]
    \begin{center}
    \renewcommand{\arraystretch}{1.2} 
    \begin{tabular}{cp{10cm}}
        \toprule
        \textbf{Prompt Name} & \textbf{System Prompt} \\
        \midrule
        Judge & You are a judge for the New Yorker cartoon caption contest\\ 
        \hdashline
        Male Lawyer & Imagine you are a white male lawyer in your 50s. You grew up in New York City and have been reading the New Yorker Magazine ever since.\\ 
        \hdashline
        Female Lawyer & Imagine you are a white female lawyer in your 50s. You grew up in New York City and have been reading the New Yorker Magazine ever since.\\ 
        \hdashline
        CS PhD & Imagine you are a computer science PhD student. You have been submitting captions for every New Yorker cartoon caption contest for the past three years.\\ 
        \hdashline
        Sociologist \& Psychologist & Imagine you are a sociology and psychology researcher that studies the New Yorker humor.\\ 
        \hdashline
        Literature Student & Imagine you are an English literature student that loves the New Yorker Magazine and its humor.\\ 
        \hdashline
        Bob Mankoff & Imagine you are Bob Mankoff, the editor of the New Yorker Cartoon Contest.\\ 
        \hdashline
        Larry Wood & Imagine you are Larry Wood, the 8-time New Yorker Cartoon Contest winner.\\ 
        \hdashline
        Cartoon Author & Imagine you are a cartoon author who often reads the New Yorker Cartoon Contest for inspiration.\\
        \bottomrule
    \end{tabular}
    \caption{Persona prompt names and their corresponding text.}
    \label{tab:system_prompt_description}
    \end{center}
\end{table*}

\section{Persona Prompt} \label{apx:system_prompt_details}
We develop different system prompts, trying to represent different demographic groups of the New Yorker Cartoon Contest audience. See Table~\ref{tab:system_prompt_description} for details.

\newpage

\section{Human Expert Annotation}\label{apx:human_stats}
To evaluate human performance, we collect assessments from five expert annotators and compute three different accuracy metrics. Each expert is given the following instruction at the beginning of the task.
\begin{center}
\fbox{
\begin{minipage}{.95\linewidth}
In each trial of this task, you will see one cartoon and two captions: the cartoon is on top, and the two caption choices are beneath the cartoon.

For each trial, \textbf{please select the caption that is the funniest for the cartoon.}

There will be around 100 trials. You will have opportunities to take a break throughout. There are attention checks during the experiment. Please chose the same image as the one on top for these trials.

Click 'Continue' to begin the test.
\end{minipage}
}
\end{center}

After the instruction page, the participants complete 100 trials, each of which looks like the following. 
\begin{center}
\fbox{
\begin{minipage}{0.95\textwidth}
    \centering
    \includegraphics[width=0.3\textwidth]{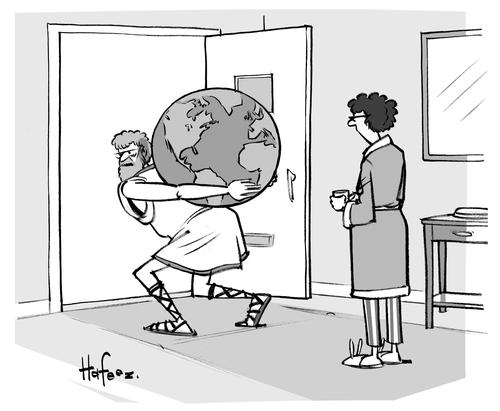}\\[1em]
    
    \begin{minipage}{0.45\textwidth}
        \centering
        Unless the apocalypse comes, I'll be back for dinner.
    \end{minipage}
    \hfill
    \begin{minipage}{0.45\textwidth}
        \centering
        This was easier to carry when it was flat.
    \end{minipage}
\end{minipage}}
\end{center}

As shown in Table ~\ref{tab:human_accuracy}, the average accuracy represents the mean performance across all five experts. For the highest accuracy metric, we independently identify the best-performing expert for each of our two ranking tasks (Rank 10 vs 1000 and Rank 30 vs 300 pairs). The majority vote accuracy reflects the performance of collective human judgment. For each test instance, we aggregate the five individual expert annotations through majority voting to determine the final prediction, then calculate the accuracy of these consensus-based decisions. To assess inter-annotator agreement, we employ two complementary metrics, Fleis Kappa and agreement rate. The Fleiss Kappa values indicate fair to moderate agreement, accounting for the chance agreement. The agreement rate measure if randomly selected two annotators' judgments for a random caption pair, they would agree 77.28\% of the time for the ranking tasks of Rank 10 vs 1000 and 67.68\% of the time for the ranking tasks of Rank 30 vs 300.

\begin{table}[t]
    \begin{center}
        \setlength{\tabcolsep}{4pt} 
        \begin{tabular}{lcc}
            \toprule
            Metrics & 10vs1000 & 30vs300 \\ 
            \midrule
            Average Accuracy & 78   & 61.6 \\
            Highest Accuracy & 85.33   & 68   \\
            Majority Accuracy & 84   & 66   \\
            Fleis Kappa       & 0.3641 & 0.2304 \\
            Agreement Rate    & 77.28 & 67.68 \\
            \bottomrule
        \end{tabular}
        \caption{Human expert performance. There are total of 5 human expert in this group.}
        \label{tab:human_accuracy}
    \end{center}
\end{table}

\section{Additional Paper Details}
We used OpenAI, Anthropic and Google APIs for all experiments. Overall, our experiments cost around \$4,000 USD. In addition, LLMs have been used to rephrase some parts of this paper.

This paper is for research purpose only, and complies with the CC-BY-4.0 license for the dataset from \citet{hessel-etal-2023-androids} and the CC-BY-NC-4.0 license from \citet{zhang2024humor}.

\end{document}